\documentclass[letterpaper, 10 pt, conference]{ieeeconf}  

\IEEEoverridecommandlockouts 

\usepackage[utf8]{inputenc}
\usepackage{booktabs}
\usepackage{balance}
\newcommand{\mpara}[1]{{\bf #1}}
\usepackage{todonotes}
\usepackage{amssymb, amsmath}
\usepackage{hyperref}
\usepackage{lipsum}
\usepackage{algorithm}
\usepackage{algpseudocode}

\title{\LARGE \bf
An Adaptive Clustering Approach for Accident Prediction
}
\author{Rajjat Dadwal$^{1}$, Thorben Funke$^{1}$, Elena Demidova$^{2}$%

\thanks{$^{1}$L3S Research Center, Leibniz University Hannover, Appelstraße 9a, 30167 Hannover, Germany 
        {\tt\small dadwal@L3S.de, tfunke@L3S.de}}%
\thanks{$^{2}$Data Science \& Intelligent Systems (DSIS) Research Group, University of Bonn, Friedrich-Hirzebruch-Allee 5, 53115 Bonn, Germany
        {\tt\small demidova@cs.uni-bonn.de}}%
}

\newcommand{\approach}{\textit{ACAP}}

\newcommand{\nnumber}{\mathbb{N}}
\newcommand{\real}{\mathbb{R}}
\newcommand{\events}{\mathbf{E}}
\newcommand{\event}{E}

\newcommand\copyrighttext{%
  \footnotesize \textcopyright 2021 IEEE.  Personal use of this material is permitted.  Permission from IEEE must be obtained for all other uses, in any current or future media, including reprinting/republishing this material for advertising or promotional purposes, creating new collective works, for resale or redistribution to servers or lists, or reuse of any copyrighted component of this work in other works.}
\newcommand\copyrightnotice{%
\begin{tikzpicture}[remember picture,overlay]
\node[anchor=south,yshift=10pt] at (current page.south) {\fbox{\parbox{\dimexpr\textwidth-\fboxsep-\fboxrule\relax}{\copyrighttext}}};
\end{tikzpicture}%
}

\begin{document}

\maketitle
\copyrightnotice
\begin{abstract}
Traffic accident prediction is a crucial task in the mobility domain. 
State-of-the-art accident prediction approaches are based on static and uniform grid-based geospatial aggregations, limiting their capability for fine-grained predictions. This property becomes particularly problematic in more complex regions such as city centers. In such regions, a grid cell can contain subregions with different properties; furthermore, an actual accident-prone region can be split across grid cells arbitrarily. This paper proposes Adaptive Clustering Accident Prediction (ACAP) - a novel accident prediction method based on a grid growing algorithm. ACAP applies adaptive clustering to the observed geospatial accident distribution and performs embeddings of temporal, accident-related, and regional features to increase prediction accuracy. 
We demonstrate the effectiveness of the proposed ACAP method using open real-world accident datasets from three cities in Germany. 
We demonstrate that ACAP improves the accident prediction performance for complex regions by 2-3 percent points in F1-score by adapting the geospatial aggregation to the distribution of the underlying spatio-temporal events. 
Our grid growing approach outperforms the clustering-based baselines by four percent points in terms of F1-score on average. 


\end{abstract}
\section{Introduction}

Prediction of traffic accidents is an important research area in the mobility, urban safety, and city planning domains. 
Such prediction is particularly challenging due to the data sparsity, the complexity of the spatio-temporal event distribution, the variety of the involved influence factors, and the complexity of their relationships.

State-of-the-art accident prediction methods (e.g., ~\cite{yuan2018hetero},~\cite{moosavi2019accident}) mainly focus on two prediction aspects, namely feature selection to identify relevant influence factors and the definition of the predictive model architecture. 
One crucial aspect, typically neglected by the existing works, is the geospatial aggregation underlying predictive models. 
Whereas some urban areas, such as city centers, have a more complex structure and tend to attract more accidents, other areas are less accident-prone. 
Hence, differently from existing works, we include geospatial aggregation as an essential factor in our modeling.
Overall, we consider the spatio-temporal accident prediction problem according to the three dimensions: geospatial aggregation, feature selection, and predictive model architecture.

The forecasting of spatio-temporal accidents is particularly challenging due to data sparsity. 
Existing works address the data sparsity by adopting coarse geospatial aggregations, such as fixed grids~\cite{chen2016learning} or entire administrative districts~\cite{el2009accident}, as prediction targets. However, neither predefined grid cells nor administrative districts adequately fit the spatio-temporal distribution of the observed events. 
Furthermore, existing works on traffic accident prediction usually consider accident datasets in US cities (e.g.,~\cite{yuan2018hetero},~\cite{moosavi2019accident}). These cities exhibit a grid-like structure, whereas European cities have the least grid-like structure~\cite{boeing2019urban}, such that the models developed for the US cities are not directly applicable to Europe.

In this paper, we propose Adaptive Clustering Accident Prediction (\approach{}) -- a novel approach to infer adaptive grids from the observed sparse spatio-temporal event distributions. We perform predictions on adaptive task-specific regions obtained through the proposed clustering-based grid growing method. As a predictive model, we rely on a neural network approach. We combine time series forecasting, in the form of Gated Recurrent Units (GRUs), with an embedding of static regional features. 
Through experiments on real-world datasets, we demonstrate that the proposed method increases the prediction accuracy compared to the state-of-the-art baselines based on fixed grids. As our experiments demonstrate, our Adaptive Clustering Accident Prediction approach outperforms several machine learning and neural network baselines regarding F1-score on the accident prediction task in several cities in Germany.

We observed that most existing works focused on evaluating the model performance based on private datasets (e.g.,~\cite{chen2016learning}), which makes them difficult to reproduce and to extend by other researchers. 
We aim to foster reproducibility, reuse, and extensibility of our work by the research community. Hence, we use only publicly available open datasets as a basis for feature extraction. 
For example, we collect the regional attributes, such as street types or the number of junctions in a region, from OpenStreetMap (OSM)\footnote{OpenStreetMap: \url{https://www.openstreetmap.org/}} - the largest publicly available source of map data. Furthermore, we build our accident prediction model on the ``German Accident Atlas''\footnote{Accident data: \url{https://unfallatlas.statistikportal.de}} -- a publicly available official dataset containing traffic accident data for Germany. 
Moreover, we make our data processing pipeline available open-source\footnote{Software: \url{https://github.com/Rajjat/ACAP}}.

\begin{figure*}[tbp]
   \centering
     \includegraphics[width=17cm, height=6cm]{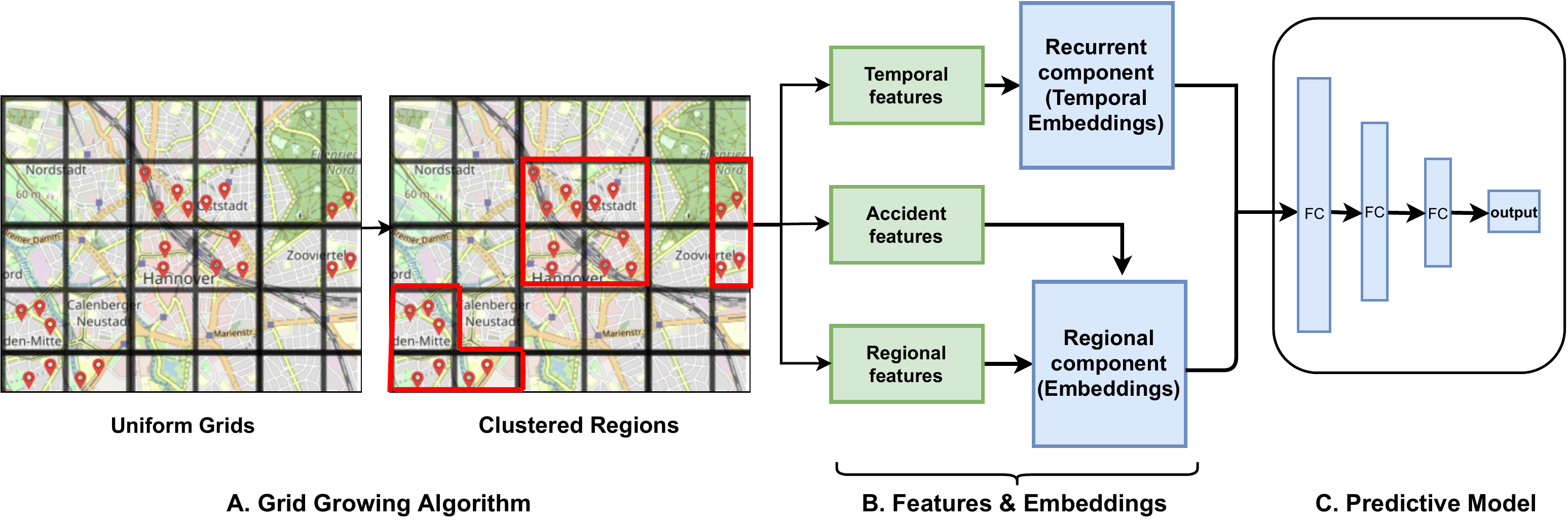}
    \caption{Architecture of the proposed \approach{} approach: a) Adaptive Clustering b) Features + Embeddings c) Classification/Prediction}
    \label{fig:approach33}
\end{figure*}


In summary, our contributions are as follows:
\begin{enumerate}
\item We propose \approach{} -- a novel approach to infer adaptive grids from sparse spatio-temporal accident distributions.
    \item Our proposed prediction model using \approach{} as geospatial aggregation achieves state-of-the-art prediction performance on the general task of traffic accident prediction and significantly improves the prediction results in the more complex areas such as city centers. 
        \item Our \approach{} approach relies on open data and an open-source pipeline. 
    \item Our experiments demonstrate that \approach{} outperforms several baselines with a performance increase of 2-3\% on average concerning F1-score on three large German cities.
\end{enumerate}

The rest of the paper is structured as follows:
First, we discuss related work in Section \ref{sec:related_work}.
Then, in Section~\ref{sec:problem} we present the formal problem statement for sparse spatio-temporal event prediction.
In Section~\ref{sec:approach}, we present our proposed \approach{} approach based on adaptive clustering with grid growing.
Section~\ref{sec:setup} describes our experimental setup, including baselines and datasets. 
We present the evaluation results on open real-world datasets in Section~\ref{sec:evaluation}.
Finally, we provide a conclusion in Section \ref{sec:conclusion}.

\section{Related Work}
\label{sec:related_work}
In this section, we discuss related work on accident prediction.
While existing approaches perform accident prediction on fixed grids or specific highways/streets, the proposed \approach{} approach adapts to the specific regions. In the following, we discuss relevant accident prediction approaches according to the spatial aggregations they adopt.

\mpara{Prediction on fixed map grids.}
Moosavi \emph{et al.} \cite{moosavi2019accident} developed a DAP model for predicting the occurrences of an accident on the 5x5 grid in a 15-minute interval. 
They evaluated the model on sparse data by augmenting it with the Point of Interests (POIs), weather, and time.
Hetero-ConvoLSTM \cite{yuan2018hetero} predicted the number of accidents on the 5x5 grid during each time slot (a day).
They used heterogeneous data, including roads, weather, time, traffic, and satellite images. Ren \emph{et al.} \cite{ren2018deep} employed an LSTM model that predicts the frequency of accidents, given the history of the past 100 hours, for 1x1 grids.
In another study by Chen \emph{et al.} \cite{chen2016learning}, the accident prediction is performed on a $500$m$\times500$m grid cell with the human mobility data as well as a set of 300,000 accident records in Tokyo (Japan). The authors predicted the possibility of accident occurrence on an hourly basis.

\mpara{Prediction on street segments.}
The works in this category deal with predicting an accident or accident count on a given road/highway. 
Chang \emph{et al.} \cite{chang2005analysis} used information such as road geometry,
annual average daily traffic and weather data to predict the frequency of accidents for a highway in Taiwan using a neural network and compared the results with the Poisson or negative binomial regression.
Caliendo \emph{et al.} \cite{caliendo2007crash} embedded road attributes such as length, curvature, annual average daily traffic, sight distance, side friction coefficient, longitudinal slope, and the presence of a junction to predict the accident count on a four-lane median-divided Italian freeway.
There are similar works related to accident prediction on highways. 
For example, an accident prediction model by Wenqi \emph{et al.} \cite{wenqi2017model} based on a convolution neural network is designed to forecast an accident on the I-15 USA highway. 
Yuan \emph{et al.} \cite{yuan2017predicting} predicted the accident occurrence for each road segment in the state of Iowa each hour in similar work. 
Hollenstein \emph{et al.} \cite{hollenstein2019investigating} investigated the association of bicycle accident occurrence to roundabout properties of the road at Swiss roundabouts using a logistic regression approach. The authors also studied various features of roundabouts responsible for bicycle accidents. 

In summary, existing approaches rely on a fixed grid of arbitrary size or a pre-defined street-segment aggregation. In contrast, \approach{} is a novel adaptive approach for predicting accidents in spatially closed regions, irrespective of the fixed grids or specific street segments. Furthermore, \approach{} works on sparse data, publicly available and easy to collect, in contrast to the approaches that use extensive but often closed datasets for modeling and prediction.

\section{Problem Statement}
\label{sec:problem}

We phrase our considered problem of traffic accident prediction in a general fashion of sparse spatio-temporal event prediction. 
Since in this paper we are only interested in predicting traffic accidents, as a particular case of spatio-temporal events, we use events and accidents as synonyms. 

Let $\events\subset\real^3$ be the set of spatio-temporal events, i.e., each event $\event\in\events$ consists of the latitude, longitude, and time information.
We are interested in the prediction of these events for different spatial aggregations:
Let $f\colon \real^3 \to \nnumber^2$ be the aggregation function mapped into $R_{\max}$ cells and $T_{\max}$ time intervals, i.e., $f(\events) \subset \{0,\dots, R_{\max}\}\times\{0, \dots, T_{\max}\}$.
We are especially interested in studying the effect of different geospatial aggregations on prediction performance.

Since time and position do not provide sufficient information for developing predictive models in this domain, we assume additional features about each spatial cell and time interval.
Formally, let $X_{\text{temporal}} \subset \real^{R_{\max}\times T_{\max}\times d_t}$ and $X_{\text{cells}} \subset \real^{R_{\max}\times d_r}$ be the matrices of the $d_t$ temporal and $d_r$ spatial features.
For example, we have as part of the regional information $X_{\text{cells}}$ the number of junctions, the street length, and the region size.
Examples of region-specific temporal features $X_{\text{temporal}}$ are solar elevation and solar azimuth.

Our task is to create a binary forecast based on k-historic observations, i.e., to train a function $\Phi \colon \{0,\dots, R_{\max}\} \times \real^{d_r} \times \real^{k\times d_t}\to \{0,1\}$ such that $\Phi$ outputs 1, if an event is observed in the next time period in the specific region, and 0 otherwise. 
We assume an imbalanced event set, where the occurrence of one event, e.g., non-accident, is much more likely than the other kind of event, e.g., accident. 
Furthermore, we are interested in comparing the performance over different spatial aggregations. 
Hence, it leads to change of the aggregation function $f\colon \real^3 \to \nnumber^2$ to another aggregation function $\Tilde{f}\colon \real^3 \to \nnumber^2$. 

\section{Approach}
\label{sec:approach}
This section presents the Adaptive Clustering Accident Prediction (\approach{}) approach proposed in this paper. 
The model architecture of \approach{} is illustrated in Fig.~\ref{fig:approach33}. 
First, we propose an adaptive clustering technique to build clusters that reflect the geospatial distribution of the accidents, presented in Section~\ref{sec:Adaptive Clustering}. 
Then, our method generates temporal and geospatial feature embeddings, presented in Section~\ref{sec:Features}.
Finally, we describe the predictive model of \approach{} in Section~\ref{sec:Prediction}.

\subsection{Adaptive Clustering with Grid Growing}
\label{sec:Adaptive Clustering}

Existing accident prediction approaches apply either a uniform geospatial aggregation using standard methods, such as geohash~\cite{morton1966computer}, or utilize administrative districts as a prediction target. The geospatial aggregation adopted by these approaches is often enforced by the already aggregated raw data, e.g., resulting from anonymization. 
The resulting uniform spatial grids are relatively coarse and do not reflect the actual accident distribution. 
Furthermore, existing works typically utilize US datasets such as Large-Scale Traffic and Weather Events Dataset (LSTW)\footnote{\url{https://smoosavi.org/datasets/lstw}}, and IOWADOT data\footnote{\url{https://public-iowadot.opendata.arcgis.com/datasets/crash-data}} for the evaluation. In these datasets, the uniform grid structure appears meaningful, as it follows the typical layout of the US cities. In contrast, the European cities' road layout does not typically follow the grid-like structure~\cite{boeing2019urban}. 
These observations motivate us to perform adaptive clustering to create geospatial aggregations that better fit the road layout and city infrastructure in the target region.

Algorithm~\ref{alg:gg} presents an overview of the adaptive clustering approach proposed in this work. This algorithm is based on our variant of grid growing~\cite{zhao2015grid}, which learns geospatial regions based on the training data, e.g., past observed accidents. 
The algorithm includes two main steps: 1) grid construction and 2) grid growing. The grid construction step requires an initial geospatial grid as a basis. 
This grid is then aggregated iteratively to form larger regions that follow the event distribution. 
In the grid growing approach proposed by~\cite{zhao2015grid}, the initial number of rows and columns is user-defined, and these parameters are not intuitive. 
In contrast, we construct the grid in a novel way with the help of geohash. 
Geohash encodes a geographic location into a string of letters and digits. Each character in the geohash defines a specific grid, e.g., ``u1qcvmz82kw'' stands for Hannover city center. 
Longer geohash values correspond to the fine-granular grids with smaller cell sizes.
In this work, we experiment with the geohash of length five, six, and seven, which approximately correspond to the regions of $4.89$km$\times4.89$km (5x5), $1.22$km$\times0.61$km (1x1), and $153$m$\times153$m (0.1x0.1), respectively. 
We experimentally assess the influence of the geohash length and utilize the geohash of length seven, which corresponds to the smaller cell size, i.e., 0.1x0.1 ($\delta_{\text{detail}}$), in our grid growing approach.

The next step is the grid growing.
In the first step, we randomly select a seed, i.e., a grid cell containing an accident. The region starts growing from the current seed by searching for accidents in the neighbor cells. 
As the eight-neighbors search gives more accurate results than the four-neighbors search~\cite{zhao2015grid}, we perform an eight-neighbors search to obtain nearby accidents in all adjacent grid cells.
The grid growing stops when the current region does not find any accidents in the adjacent grid cells and assigns a cluster to the resulting region. 
In the next step, we choose the next seed cell randomly from the accident-prone grid cells not clustered in the previous algorithm iterations.
The grid growing algorithm continues until it assigns all accident-prone grid cells in the training set to a cluster. 
Based on the clusters generated by the grid growing algorithm, we can, later on, assign 
locations and accidents unseen during training to their nearest clusters.
To define the nearest cluster, we adopt haversine distance and apply a distance threshold $\Delta$.
We experimentally set $\Delta$ = 400 meters. 
For the accident locations not mapped to any of the clusters due to the distance value exceeding the threshold, we map those locations to a larger base grid cell of 1x1 ($\delta_{\text{base}}$) and assign this cell to a separate geospatial cluster. 

\begin{algorithm}[htbp]
\caption{Adaptive Clustering with Grid Growing}
\label{alg:gg}
\small
\begin{algorithmic}[1]
\State \textbf{Input:} Spatio-(temporal) events $\events$, e.g., training set of accidents
\State \textbf{Output:} Spatial-aggregation function $f_{\text{GG}}$
\State \textbf{Hyperparameters:} detailed grid size $\delta_{\text{detail}}$, base grid $\delta_{\text{base}}$, distance threshold $\Delta$
\State Calculate for each $\event \in \events$ their detailed grid $G^{\delta_{\text{detail}}}(\event)$
\State Initialize clusterings $\mathbf{C}= \emptyset$ and $i=0$
\While{Unmarked event $\event\in \events$ exist}
\State Select random unmarked event $\event \in \events$
\State Set $C_i=\{\event\}$
\Repeat
\State Check for each event in $C_i$ the 8-$G^{\delta_{\text{detail}}}$-neighborhood for events $\events_\text{neighbors}$
\State Set $C_i= C_i \cup \events_\text{neighbors}$
\Until{No new neighbors, i.e., $\events_\text{neighbors}=\emptyset$}
\State Mark all events in $C_i$ and set $\mathbf{C}= \{C_0, \dots C_i\}$
\EndWhile
\State \textbf{return} $f_{\text{GG}}(\event) =
\begin{cases} C, &\text{if } C=\operatorname{argmin}_{\Tilde{C} \in \mathbf{C}} \operatorname{d}(\event, \Tilde{C}), \\
 & \text{and } \operatorname{d}(\event, C)<\Delta,\\ 
G^{\delta_{\text{base}}}(\event) & \text{otherwise} 
\end{cases}$
\end{algorithmic}
\end{algorithm}

The grid growing algorithm illustrated in Fig.~\ref{fig:approach33} is essential for building adaptive regions. 
We compare the proposed grid growing approach to fixed grids and clustering approaches in the evaluation.
The advantages of adaptive clustering, and especially of the grid growing approach proposed in this work, are as follows:
(i) Our geospatial aggregation adapts to the underlying distribution of accidents in the dataset. In other words, we adjust the geospatial resolution based on the events that occur in the geospatial proximity. 
(ii) Our adaptive clustering allows us to work with sparse spatio-temporal data, unlike other baselines~\cite{moosavi2019accident}. This property makes our approach easily applicable to large (rural) areas where the data can be extremely sparse. 

\subsection{Features \& Embeddings}
\label{sec:Features}

As a data pre-processing step, we compute temporal and geospatial features such as accident and regional features for each adaptive cluster and each grid cell. 
We evaluate the adaptive clustering approach with a fixed grid of cell size 5x5 and 1x1 in Section \ref{sec:evaluation}.

\mpara{Temporal Features.}
Accidents are time-dependent, such that we aim to learn the correlation between the accidents and the temporal features.
Our model includes ten temporal features such as weekday/weekend, season, month, year, weekdays, an hour of the day, daylight, solar position, solar azimuth, and solar elevation. 
All temporal features are encoded in one feature vector using the one-hot-encoding technique. The resulting feature vector includes 36 dimensions, where each dimension represents a possible feature value. 
The degree of temporal aggregation depends, in general, on data availability. 
In the ``German Accident Atlas'' dataset used in the evaluation, temporal features are aggregated on an hourly basis due to legal restrictions.

\mpara{Accident Features.}
The accident features include the accident type and the road conditions during the accident. 
Examples of accident types in the ``German Accident Atlas'' dataset include a car collision with another car or a bicycle. 
A specific accident type can be more prominent at one location than others, e.g., a city center has more car collisions than collisions with a bicycle.
Thus, the accident type feature helps to identify such areas. 
Road conditions feature informs whether the road was wet, slippery, or dry during an accident. 
The accident features are converted into one-hot-encoded vectors and averaged for the accidents in a geospatial cluster or a grid cell.

\mpara{Regional Features.}
Regional features are infrastructural attributes of a specific region, i.e., a grid cell or an adaptive cluster.
Intuitively, regional features have a significant influence on accident occurrences. 
For example, accidents tend to occur more often near junctions or crossings. 
We select the following Point of Interests (POIs) as regional features: 
amenities count, number of crossings, number of junctions, number of railways, station frequency, stop signs count, number of traffic signals, number of turning loops, number of giveaways, highway types, and the average maximum speed for each region.
We normalize feature values to the range between 0 and 1. 
We extract regional features from OSM.

\mpara{Feature Embedding.}
Embeddings are continuous vector representations of discrete variables. 
Embeddings can help to reduce the dimensionality of feature vectors and to represent latent features. 
We construct latent representations from one-hot-encoded and normalized feature vectors generated above as follows. 

\textbf{Temporal embeddings.}
For the temporal features, we utilize Gated Recurrent Unit (GRU) to create temporal embeddings. 
GRU is a type of Recurrent Neural Network (RNN) to learn sequential or temporal data. 

A set of eight temporally ordered one-hot-encoded vectors from the preceding time points, each of length $n$, where $n$ corresponds to the number of one-hot-encoded features, are fed to the GRU. With the temporal features listed above, $n$=36.
GRU includes two recurrent layers in our settings, each with 128 units, and outputs the embedding vector of the same length.

\textbf{Embeddings of accident and regional features.}
For these features, a feed-forward layer of size 128 with the sigmoid activation function creates feature embeddings.

\subsection{Predictive Model}
\label{sec:Prediction}

The predictive model of \approach{} outputs a softmax, i.e., the likelihood for accidents respectively non-accidents. 
We transform them into binary accident labels, i.e., '1' for accident and '0' for non-accident.
The model input is composed of the temporal embeddings and the embeddings of the accident and regional features of each geospatial cluster or grid cell.
The input is feed-forwarded through the neural network layers with decreasing dimensionality. In particular, we use a set of fully connected layers of size 512, 256, 64, and 2, respectively. The activation function is applied in each layer to induce non-linearity in the model.
The first three layers utilize ReLU as the activation function, whereas we apply softmax activation to the last layer's output. 
We use batch normalization~\cite{ioffe2015batch} after the second and third layers. The role of batch normalization is to re-scale and normalize the intermediate outputs. The last layer is the classification layer that predicts binary accident labels. We optimize \approach{} using categorical cross-entropy as a loss function.

\section{Evaluation Setup}
\label{sec:setup}
In this section we describe the baselines, datasets, parameters and metrics utilized in the evaluation.

\subsection{Accident Prediction Baselines}

We utilize four baseline methods, including machine learning and deep learning baselines: \textit{Logistic Regression (LR), Gradient Boosting Classifier (GBC), Deep Neural Network (DNN), and Deep Accident Prediction (DAP)} model \cite{moosavi2019accident} to compare the performance of our approach regarding accident prediction.

LR is widely used for classification tasks where the model outputs probabilities for classification problems. GBC, another ML-based baseline with boosting characteristics, is also suitable for our classification task. 

To compare our approach with deep learning models, we use DAP and DNN. DAP utilizes Long Short-Term Memory (LSTM) for temporal learning, Glove2Vec for learning accident descriptions, and embedding components for learning spatial attributes. DNN employs a set of fully connected layers of size 512, 256, 64, and 2, respectively.

\subsection{Clustering Baselines}

Geospatial aggregation can be broadly divided into two parts: grid-based and clustering-based. 
For the grid-based aggregation, we use the 5x5 and more detailed 1x1 geohash grids, as described in Section~\ref{sec:Adaptive Clustering}.
The clustering approaches belong to the three categories: neural network-based, density-based, and centroid-based.
As a representative of the neural network-based clustering methods, we evaluated Self-Organizing Map (SOM)~\cite{kohonen1996engineering,mangiameli1996comparison,846731}.
In density-based clustering, DBSCAN~\cite{ester1996density} and its extension Hierarchical DBSCAN (HDBSCAN) have been used to cluster the geospatial data \cite{zhao2015grid}.
DBSCAN is an unsupervised machine learning algorithm to classify unlabeled data. 
As a representative of the centroid-based methods, we apply the well-known K-means algorithm~\cite{macqueen1967some}.

\subsection{Dataset}
\label{sec:dataset}

The accident dataset is collected by ``The Federal Statistical Office'' department in Germany and is openly accessible. 
This dataset includes accident information for 16 German federal states starting from 2016 and currently contains data until 2019. 
The dataset contains 24 accident attributes, including accident id, latitude, longitude, day of the week, hour, month, year, accident type, and road condition.
Due to Germany's legal restrictions, the data is aggregated temporally on an hourly basis, and the specific date of the accident is not reported in the dataset.
We filtered the dataset to obtain cities with a long observation period and a sufficient number of accidents to facilitate model training and selected Hannover, Munich, and Nuremberg. 
For example, Hannover and Nuremberg have comparable accidents count with 7,433 and 6,121, respectively. In contrast, Munich accounts for the highest number of accidents, with 14,986 accidents in the considered period.

\mpara{OpenStreetMap Dataset.}
OSM is a publicly available geospatial database. One can easily extract and store regional features such as POIs from OSM geofabrik\footnote{\url{https://download.geofabrik.de/europe/}}.
For example, around 50 percent of the accidents happened at primary, secondary, tertiary, and trunk highways in Lower Saxony, Germany. 
The aim is to leverage our model with regional features to help in the prediction task.
We fetch the regional features from the OSM dataset, e.g., number of amenities, number of junctions, number of traffic signals, 
and different highway types. 
We aggregate each regional feature to its 0.1x0.1 geohash and map it to the clusters and grids in our settings.

\mpara{Negative Samples.}
Accident prediction is a binary classification task that requires generating elements of the non-accident class. 
Any spatio-temporal point where no accident has occurred can be considered as a non-accident. 
However, using all time points leads to the generation of too many non-accidents.
To compare different spatio-temporal aggregations on the same dataset, we randomly select a 0.1x0.1 geohash grid and randomly generate a temporal and spatial point for the selected grid. 
Motivated by~\cite{yuan2017predicting}, we maintain a fixed accident to non-accident ratio, i.e., 1:3 across training and test data. 

\mpara{Training and Test Split.}
We split three years of data into training and test data: first 29 months, i.e., 80\% of data for training, and last seven months, i.e., 20\% for testing.
For validation, we utilize the hold-out cross-validation method. 
In this method, a subset (10\%) of the training data (split temporally) is reserved for validating the model performance.
The early stopping technique based on the validation set is performed as a regularization step with patience as an argument. 
Patience represents the number of epochs before stopping once the loss starts to increase. 
We train each model separately for each city and perform testing on the same city.

\subsection{Hyperparameters}
In the following, we describe the hyperparameter settings of the models adopted in the evaluation.

\mpara{Clustering Baselines.}
We initialize the hyperparameters of the clustering baseline as follows.
DBSCAN takes epsilon $(e)$ and the minimum number of points $(n)$ as input parameters. 
The value of $e$ is determined by the DMDBSCAN algorithm~\cite{rahmah2016determination} using the nearest neighbor search. The selected $e$ with a combination of different values of $n$ help to determine silhouette scores~\cite{rousseeuw1987silhouettes}. The values of $e$ and $n$ with the highest silhouette score are chosen.
HDBSCAN has minimum cluster size as the only parameter, which we set to four.
We apply the elbow method to determine the number of clusters in K-means (K=4).
For SOM, we choose a map size of $30\times30$, which gives a comparable number of clusters as the 1x1 grid.

\mpara{Model Hyperparameters.}
We find the best parameter setting for the aforementioned ML-based baseline models by using grid-search. 
We follow the same setting as in~\cite{moosavi2019accident} and refer to our available code for further details about the baselines' hyperparameters.

For \approach{}, Adam optimizer with an initial learning rate of 0.01 is used to train the model.
A dropout of 0.2 is used for regularization in the GRU layer. In early stopping, patience with 15 helps in regularization.
For DNN, the parameter setting is the same as in the fully connected predictive model of the \approach{}. All the neural network-based models are trained for 60 epochs.

\begin{figure}[b!]
   \centering
     \includegraphics[width=.9\linewidth,height=5cm]{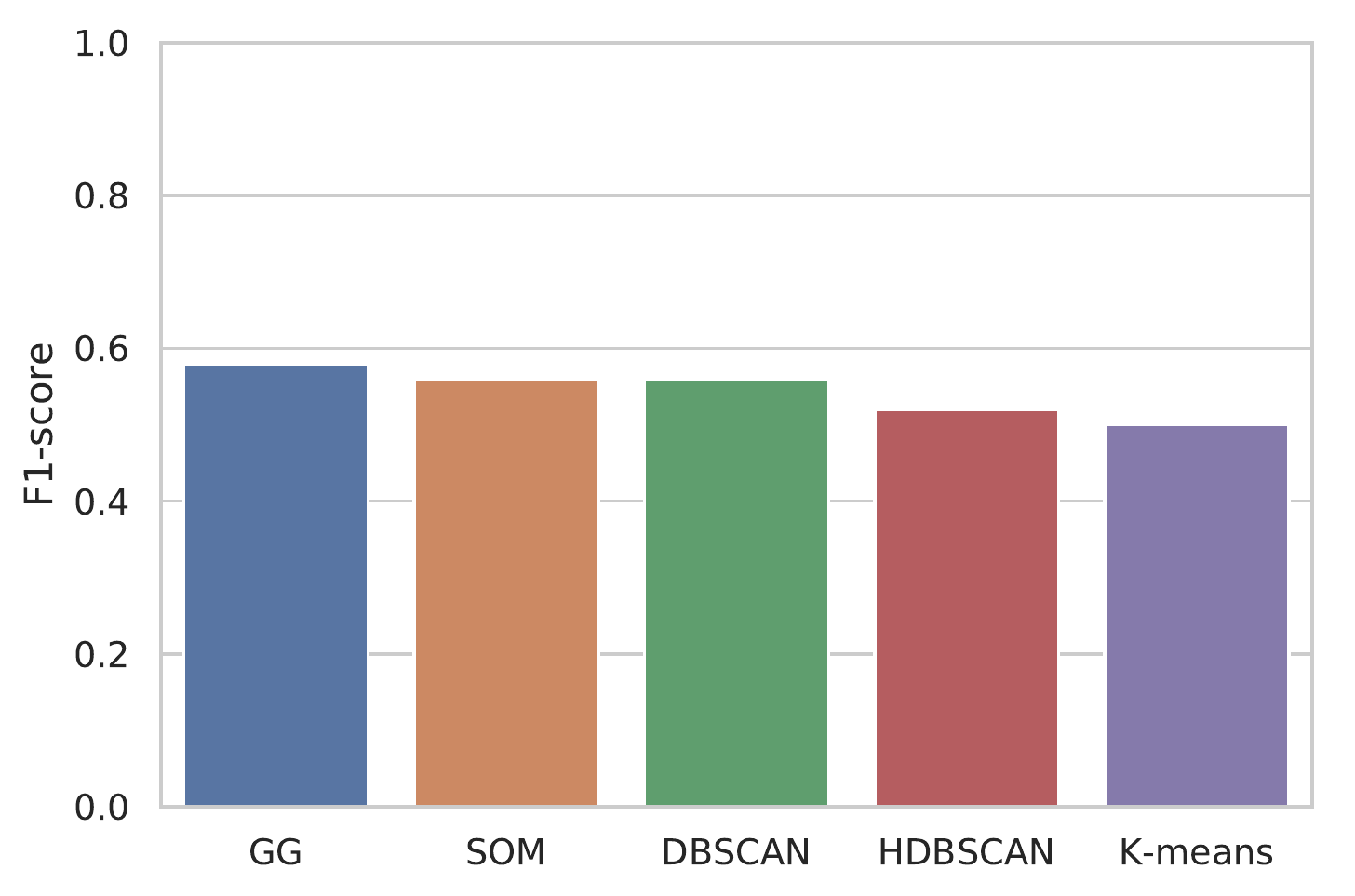}
    \caption{Comparison of spatial clustering methods for Hannover city (\approach{})}
    \label{fig:approach3}
\end{figure}

\subsection{Evaluation Metric}
Due to uneven class distribution, we use F1-score as a metric for evaluating different models. 
F1-score is the harmonic mean of precision and recall.
Since we are interested in predicting the accident class, we report the F1-score of the accident class for different models. We run each model ten times and report the average F1-score. 

\section{Evaluation}
\label{sec:evaluation}
The evaluation aims to assess the proposed accident prediction approach, analyze the effect of the proposed adaptive geospatial clustering and examine feature importance.

\subsection{Effect of Geospatial Clustering}
As the first step of the \approach{} evaluation, we compare different spatial clustering methods by changing the clustering in \approach{}. 
In other words, we change the adaptive clustering (AC) part of our approach and plug in other clustering methods. 
Fig.~\ref{fig:approach3} shows that our grid growing approach, i.e., GG outperforms all baselines by at least four percent points.
The best performing baselines are SOM and DBSCAN, while HDBSCAN and K-means result in the worst model performance. 
Overall, we can observe that the proposed geospatial clustering has a significant positive effect on the observed performance. 
To further analyze our model and the geospatial aggregation, we evaluate \approach{} and the best clustering baseline SOM against two uniform grids on three different cities in the next section.

\subsection{General Performance}
To extensively study \approach{} performance, we evaluated \approach{} using four different spatial aggregations and four other prediction methods on three German cities. 
As Table~\ref{tab:regions} shows, our \approach{} approach achieves the highest F1-score in the accident prediction for all spatial aggregations and all cities. 
With respect to spatial aggregation, our grid growing clustering and 1x1 grids achieve the best results, while especially 5x5 grids reduce the prediction quality. 
We observe that the aggregation of static features in large uniform grids negatively impacts the performance and only achieves a one percent point higher score in Hannover than K-means, while having 31 regions instead of four. 
Overall, \approach{} increases F1-score by 2-3 percent points over the best performing baseline on average. 

\subsection{Performance in the City Centers}
\begin{figure}[h!]
   \centering
     \includegraphics[width=.9\linewidth,height=5cm]{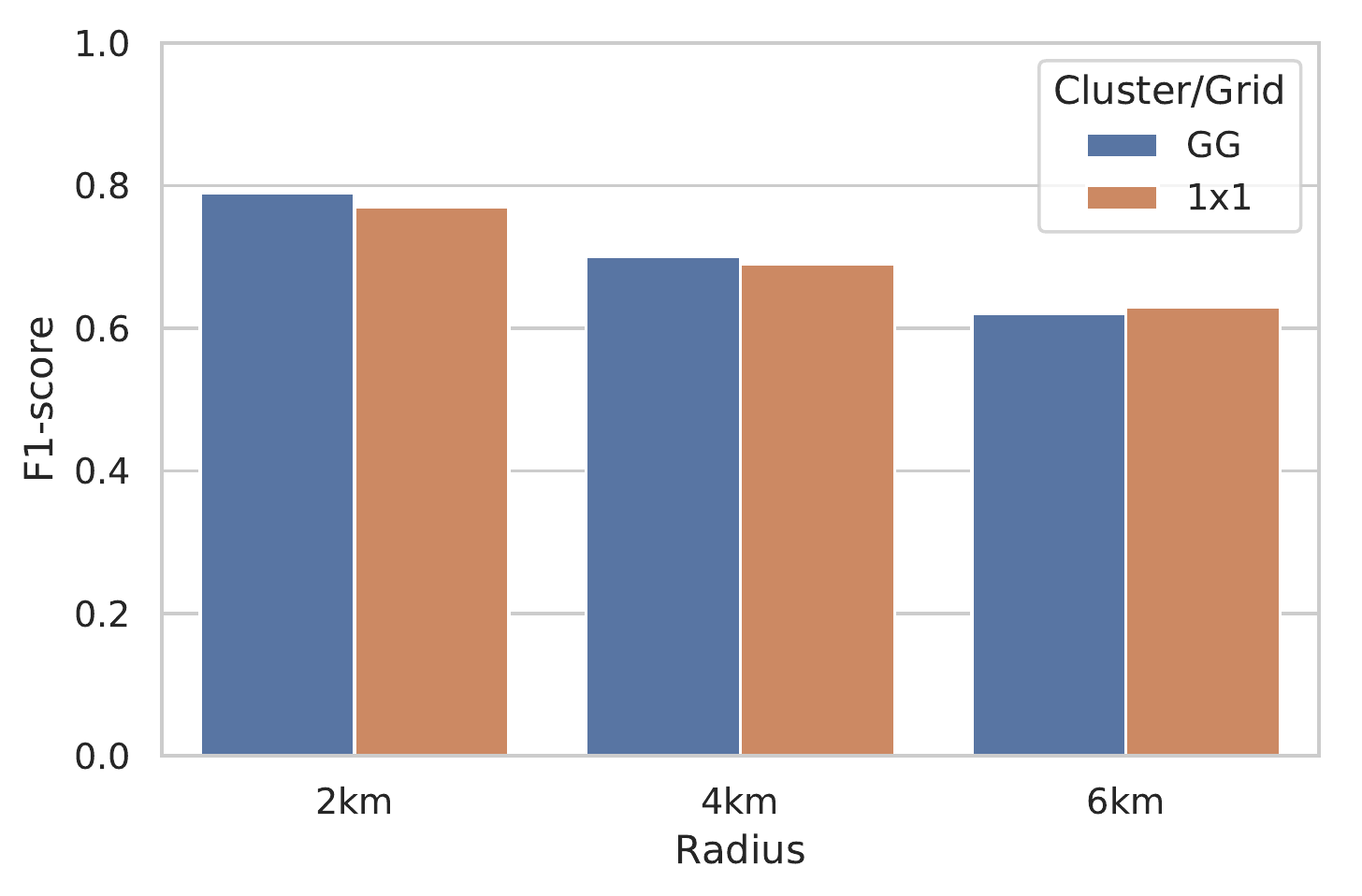}
    \caption{F1-score vs radius from the city center of Hannover (\approach{}) }
    \label{fig:radius}
\end{figure}

To further analyze the proposed grid growing algorithm in urban regions, 
we evaluate the performance of our approach starting from the city center of Hannover to the larger Hannover region. 
For simplicity, we select a different radius around the city center of Hannover and compare the performance of grid growing and 1x1 grids. 
As Fig.~\ref{fig:radius} illustrates, \approach{} with grid growing outperforms the uniform grids in the inner city center by 2 percent points.

\begin{table}[htbp]
    \centering
    \caption{F1-score of accident predictions of different cities with different aggregations
    }
    \small{
    \begin{tabular}{ll c c c c c c}
    \toprule
         Clustering & Method & Hannover & Munich & Nuremberg  \\
         \midrule
        Grid-Growing    & \approach{} &  \textbf{0.58}& \textbf{0.56}&\textbf{0.60}   \\
        Grid-Growing    & DAP &0.47  &0.44 &  0.47  \\
        Grid-Growing    & DNN & 0.55 &0.52 &0.54    \\
        Grid-Growing    & LR & 0.56 & 0.49&  0.53  \\
        Grid-Growing    & GBC & 0.52 & 0.52&0.56    \\
        \midrule
        SOM    & \approach{} &\textbf{0.56}  & \textbf{0.55} & \textbf{0.57}   \\
        SOM    & DAP & 0.42 &0.44 &   0.45 \\
        SOM    & DNN & 0.51 &0.51 &  0.54  \\
        SOM    & LR & 0.53 &0.46 &  0.53  \\
        SOM    & GBC &  0.52&0.51 & 0.54   \\
        \midrule
         1x1    & \approach{} & \textbf{0.59} & \textbf{0.57} & \textbf{0.60}  \\
        1x1    & DAP &  0.49& 0.49& 0.51   \\
        1x1    & DNN &  0.57& 0.52&  0.57  \\
        1x1    & LR &  0.57&0.52 &  0.57  \\
        1x1    & GBC &  0.52&0.51 & 0.55   \\
        \midrule
        5x5    & \approach{} & \textbf{0.52} &\textbf{0.51} &  \textbf{0.53} \\
        5x5    & DAP & 0.45 &0.44 &  0.45  \\
        5x5    & DNN &  0.50& 0.49&  0.51  \\
        5x5    & LR & 0.49 &0.40 & 0.48   \\
        5x5    & GBC &  0.16&0.26 & 0.18   \\
         \bottomrule
    \end{tabular}
    }
    \label{tab:regions}
\end{table}

\subsection{Feature Importance}
\begin{figure}[h!]
   \centering
     \includegraphics[width=.9\linewidth,height=5cm]{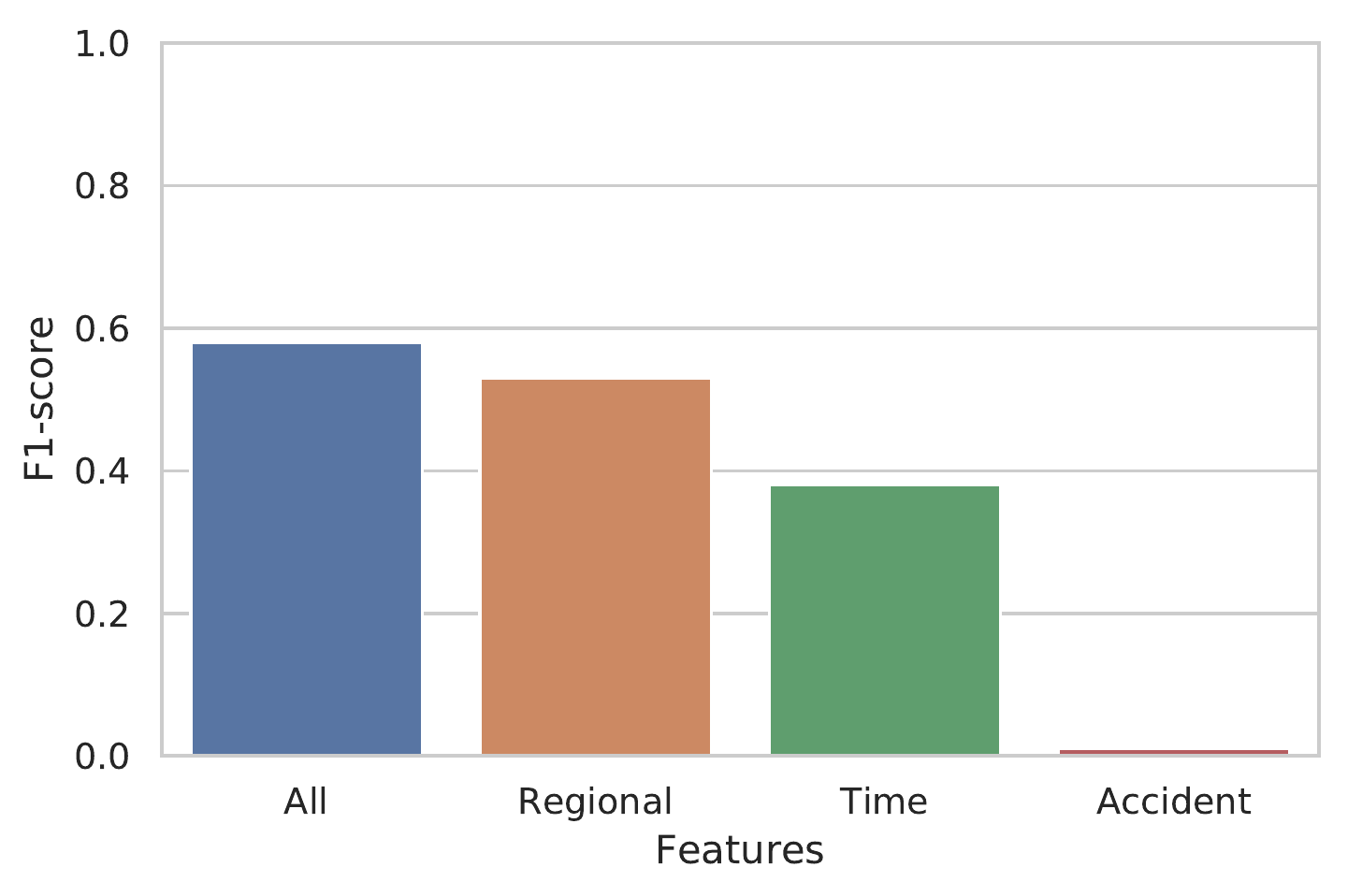}
    \caption{Effect of different features (used alone) on F1-score for Hannover (\approach{}) }
    \label{fig:Ablation}
\end{figure}

As the final part of our evaluation, we study the importance of our three feature groups -- regional, temporal, and accident features -- for \approach{}'s performance. 
Fig.~\ref{fig:Ablation} shows the resulting accident F1-score if the model only uses one feature category for the prediction. 
We observe the high relevance of regional and temporal features, which achieve 91\% and 63\% of the model that relies on all features, correspondingly. 

\section{Conclusion}
\label{sec:conclusion}

In this paper, we proposed \approach{} -- an approach that relies on novel adaptive clustering and various temporal and regional features to predict traffic accidents. Overall, we achieved a 2-3 percent points increase in F1-score over the best-performing baseline on average. Our proposed grid growing algorithm, which flexibly adapts to the regions based on the observed geospatial accident distribution, increases the performance by four percent points against the clustering-based baselines. 
We observed that our grid growing approach improves the prediction performance by two percent points in the city centers. 
Furthermore, \approach{} is based on an open data pipeline, which comes with our publicly available implementation, making the proposed approach reproducible and reusable. 
In future work, we plan to investigate the impact of user-centric features, such as driver behavior, on accident prediction.

\section*{Acknowledgements}
This work is partially funded by the BMWi, Germany under the projects ``CampaNeo'' (grant ID 01MD19007B), and ``d-E-mand'' (grant ID 01ME19009B), the European Commission (EU H2020, ``smashHit'', grant-ID 871477) and DFG, German Research Foundation (``WorldKG'', DE 2299/2-1). 

\balance
\bibliographystyle{IEEEtran}
\bibliography{ref}
\end{document}